%% file: main.tex
\documentclass[12pt]{article}
\usepackage[,a4paper, top=0.85in, left=1in, right=1in, footskip=0.75in]{geometry}

\usepackage{float}
\usepackage{tabularx}
\usepackage{xltabular}
\usepackage{makecell}
\usepackage{multirow}

\usepackage{booktabs}
\usepackage{graphicx}
\usepackage{eurosym}

\usepackage{adjustbox}

\usepackage[tableposition=top]{caption}
\usepackage{subcaption}

\usepackage[usenames,dvipsnames]{xcolor}
\usepackage{comment}

\usepackage{authblk}

\usepackage{amsmath}
\usepackage{mathtools}
\usepackage{amssymb}
\usepackage{siunitx}

\usepackage[english]{babel}
\usepackage[]{microtype}

\usepackage{setspace}
\usepackage{footnote}
\usepackage{url}
\usepackage{soul}
\usepackage{enumitem}
\usepackage{parskip}
\usepackage{longtable}

\usepackage{hyperref}
\usepackage{cleveref}

\usepackage[bitstream-charter]{mathdesign}
\usepackage[T1]{fontenc}

\usepackage{natbib}
\title{On the causality-preservation capabilities of generative modelling}
\author[1]{Yves-Cédric Bauwelinckx}
\author[1]{Jan Dhaene}
\author[2]{Milan van den Heuvel\footnote{Contact: \href{mailto:Milan.vandenHeuvel@UGent.be}{Milan.vandenHeuvel@UGent.be} (Address: Tweekerkenstraat 2, 9000, Ghent, Belgium).} 
}
\author[1,3]{Tim Verdonck}

\affil[1]{Department of Economics, Katholieke Universiteit Leuven, Belgium}
\affil[2]{\normalsize Department of Economics, Universiteit Gent, Belgium}
\affil[3]{IMEC, Universiteit Antwerpen, Belgium}
\date{}

\begin{document}

\maketitle

\include{chapters/introduction}

\include{chapters/problem_setup}
\include{chapters/generative_models}
\include{chapters/results}

\include{chapters/applications}
\include{chapters/conclusion}

\appendix

\bibliography{references}

\end{document}

%% file: chapters/introduction.tex
\section{Introduction}

To make sense of the complexities of reality, and make optimal decisions accordingly, organisations and researchers have always striven to come up with models that can accurately represent observed phenomena (e.g. consumption behaviour, loan defaults). In the past, these models were defined by the analyst and calibrated to (small) data. Recently, however, during the so-called machine learning revolution, the focus shifted to a more data-driven, algorithmic approach. Machine learning algorithms now search for the optimal model by finding support for it in the data instead of being chosen by the analyst. This approach has increased the collection of, investment in, demand for, reliance on, and value of data for organisations and research significantly~\citep{mckinseyreport}. It has also brought the tension between utility of data and privacy of its subjects to the forefront of public discussion~\citep{eu-AI-dir}. Recently developed generative modelling methods, which generate data with a distribution similar to the original but without containing any of the real data, have been proposed as a potential solution~\citep{gartnerreport,Castellanos2021}. Decision-making is, however, almost always a causal question and little is known about the replication capabilities of these methods beyond correlations. For this reason, this paper seeks to fill the gap by performing an investigation of the causal replication capabilities of data replication methods as well as defining a path forward to making them a viable option for decision-making.

There are a lot of advantages to the algorithmic approach to modelling, the most important being increased performance and the opportunity for analysts to be systematic and transparent about the process by which the model was selected~\citep{Athey2018ML}. The power of this approach has been apparent in several fields that have had incredible advances in replicating reality due to the availability of large amounts of data. One of the most famous examples is ImageNet, a database with millions of hand-labelled pictures, enabling revolutionary progress in image recognition~\citep{AlexNet}. GPT-3, a multi-purpose natural language model, similarly achieved impressive results after learning from a data set containing 45 TB of plain text~\citep{DBLP:gpt3}. Besides these topics focused around machine learning, examples can also be found in other fields such as physics and astronomy which have collected ever-growing volumes of data to learn from. Projects like the Large Hadron Collider~\citep{Evans2009} and the imaging of the black hole at the centre of galaxy M87~\citep{Castelvecchi2019} handle data in the order of petabytes. However, such large amounts of data are not always readily available. In many fields centring around individuals, such as the social and health sciences (e.g. finance, insurance, medical fields), the collecting or sharing of such datasets is far from trivial due to ethical and privacy concerns~\citep{Koenecke2020}. One recently emerging option to alleviating such concerns is generative modelling.

Generative models are models that try to learn a representation of the (high-dimensional) distribution of a dataset. Once this representation is learned, it can then be used to generate new samples that maintain the original dataset's distribution of features but that did not appear in the original dataset~\footnote{Note that to have such privacy guarantees, one needs to explicitly include an optimisation for it in the model fitting step such as in~\cite{yoon2018pategan}. Else there are cases when replication could occur~\citep{Feng2022}.}. Generative methods are thus capable of simulating non-existent but realistic-looking data, also referred to as \emph{synthetic data}, that can be shared more freely. A well-known use-case are pictures of human faces for computer vision applications. Even in the possession of a large dataset of pictures of human faces, sharing this freely could present issues concerning privacy. However, generative models are capable of constructing fake but human-looking faces that can, due to their non-existence, be shared more freely to further the quality of applications. 

While generative modelling has been around for decades, a major breakthrough in the ability to efficiently training such models was achieved in 2014 with Generative Adversarial Networks (GANs)~\citep{goodfellow2014generative}. This method increased our capacity to fit high-dimensional distributions of data, like images and video data. The GAN framework has found widespread applications throughout computer vision, like image generation~\citep{GAN_faces,GAN_images}, text to image translation~\citep{GAN_texttoimage}, the blending of images~\citep{GAN_photoblending}, enhancing quality of pictures~\citep{GAN_superres}, filling in blanks in pictures~\citep{GAN_inpainting}, and a more infamous example of deepfakes~\citep{GAN_deepfakes}. While these are noteworthy variations and applications of the GAN framework, the common factor here is the focus on computer vision. In contrast, GANs have found limited adoption within the human sciences, like economics.

The main reason for this is that in these fields, most questions are inherently about identification of causal effects. Neural networks, which are at the centre of the GAN framework, in contrast, still focus mostly on high-dimensional correlations. An example of this is shown in the paper by \cite{cow_causal}, where they analyse a neural network trained to classify images. The neural network appears to be able to accurately identify whether or not there is a cow in a picture, until you ask the network to classify a picture of a cow in an uncommon environment. The model is, for instance, not able to recognise a cow on a beach, because of the spurious correlation between cows and grasslands. Learning to label images with grass in it are shortcuts that expose the lack of generalisation of the neural network. Recently, a field has emerged called \emph{Causal Machine Learning} where researchers try to make steps towards making machine learning models more causal~\citep{Scholkopf2021}. While this field is promising, due to the inverse problem nature of finding causality in observational data, it is currently still in its infancy in regards to applicability. As we will show below.

The most prevalent used loss-functions for GANs are some form of binary cross-entropy~\citep{goodfellow2014generative,Yoon2019TimeseriesGA,GAN_images,2020Wiese} or Wasserstein distance~\citep{arjovsky2017wasserstein,CTGAN,WGAN_athey}. These losses indicate in some form or another the difference between two joint probability functions. Replicating the joint probability distribution, however, does not guarantee replication of the underlying causal process. Finding the causal structure from observational data is an inverse problem, finding the cause from the effect. Consider this example in The Black Swan from Taleb~\citep{Taleb2010}:

"\textit{Operation 1 (the melting ice cube): Imagine an ice cube and consider
how it may melt over the next two hours while you play a few rounds of
poker with your friends. Try to envision the shape of the resulting puddle.}\\\\
\textit{Operation 2 (where did the water come from?): Consider a puddle of
water on the floor. Now try to reconstruct in your mind's eye the shape of
the ice cube it may once have been. Note that the puddle may not have
necessarily originated from an ice cube.}"

Operation 1 is an example of the forward way of thinking, where the effect (the water) is to be predicted from the cause (ice cube). With the right models it is possible to accurately come up with the resulting pool of water. In contrast, operation 2 asks the inverse, finding the shape of the cube (cause) from the pool of water (effect). There are however an almost infinite amount of possible ice cubes that could have led to that pool of water. This example also translates to joint probability distributions and underlying causal models. For a given joint distribution there are a multitude of possible underlying causal models. 

In this paper, we survey the literature on generative adversarial networks, being the dominant model among generative models for synthetic data, and evaluate their capacity to preserve certain causal structures (i.e. cross-sectional, time series, and full structural) in the synthetic datasets they generate. We do so by first generating a dataset where the data-generating function, and thus the structural causal model, is know. Secondly, we make a synthetic copy of this with a specific GAN method and perform different causal analyses with an increasingly lenient set of assumptions, from cross-sectional to time-series to structural. Lastly, we check if the results in the real data align with those in the synthetic data to evaluate the causality preserving capabilities.

We find that for relationships in data where the assumptions hold such that correlation equals causation, inference on the real and synthetic data yield the same results only in the case where the actual causal structure aligns with the most simple model that can replicate the correlations in the data. In more complex cases, for instance when a variable has time-dependence and both influences cross-sectional features as well as itself, we find that the generative model converges on a model with the same general distribution, but that it does so with a simpler underlying causal structure. Our results point at the reason being the often-used regularisation in machine learning that builds in a preference for smaller models (as posited in occam's razor) which is not necessarily a valid principle in causality. Finally, when the whole causal structure is considered, it becomes apparent that currently the applicability is still limited due to the stringent assumptions that need to be met in order to overcome the challenges of the inverse problem.

The remainder of this paper is structured as follows. In Chapter 2, we lay-out the problem setup and discuss the structural approach we take to evaluate the causal replication capacity of GAN-based models. In Chapter 3, we give a general introduction to the inner workings of GAN-models and detail three different GAN variations that we take as representative for the different streams in the GAN literature that aim to capture increasingly complex correlations (i.e. cross-sectional correlations, time-series correlations, full causal structure). In Chapter 4, we present the results of our evaluation. In Chapter 5, we discuss some of the additional real-world challenges that we abstracted away from but that need to be considered where these methods to be used in real-world cases. Lastly, in Chapter 6, we summarise and conclude our findings.

%% file: chapters/problem_setup.tex
\section{Problem setup}
The goal of the evaluation in this paper is to see if current data replication methods are useful when causal analyses are to be performed on the resulting synthetic data. To simulate realistic use-cases of synthetic data for sectors/fields where decisions require causal inference, we take popular causal inference methods from the field of econometrics. These models have the benefit of having well-known sets of assumptions under which the estimated parameters may be interpreted as causal effects. First, we consider a cross-sectional model, namely ordinary least squares. This is the simplest case in which the researcher assumes that observations at different timestamps are independent and the functional form of the model is linear in the parameters. While being the simplest model to estimate in econometrics, it also comes with the most stringent assumptions to be interpreted as causal, discussed in Section~\ref{section:cross-sectional-assumptions}. Next, we look at the class of time-series models that allow dependence between observations at different timestamps in Section~\ref{section:time-series-assumptions}. Lastly, in Section~\ref{section:structural-model-assumptions}, we discuss the case where a full structural equation model is the goal of the estimation. Here, an approach from the field of \emph{Causal Discovery} is needed since none such data-driven method exists in econometrics.

\begin{figure} [H]
    \centering
    \includegraphics[width=\textwidth]{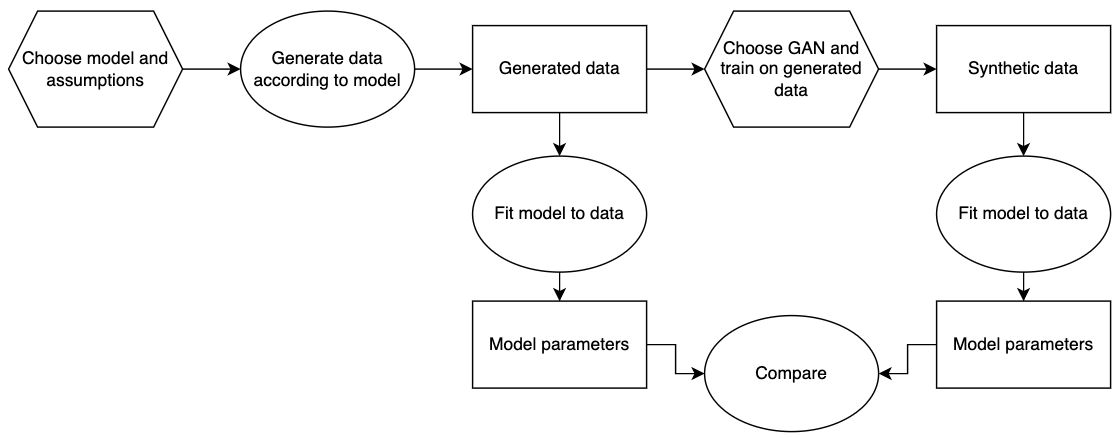}
    \caption{Experiment setup for each choice of assumptions and GAN method}
    \label{fig: experiment_setup}
\end{figure}

The evaluation setup is shown in Figure~\ref{fig: experiment_setup}. First, a dataset is generated with known causal structure. The design of this dataset is discussed in Section~\ref{sec: generated dataset}.
The generated data is then used to train a chosen type of GAN, designed to capture the distribution of the generated data, further discussed in Chapter~\ref{chapter/GAN}. The trained GAN is then sampled to construct a synthetic dataset. Note that, in the remainder of this paper, artificial data we generate from the known model are referred to as \textit{generated data} while data sampled from the GAN models are referred to as \textit{synthetic data}. Next, a causal inference method, with its accompanying assumptions, is selected to apply to both the generated data and the synthetic data. We first validate that the chosen causal inference method is appropriate to estimate (a subset of) causal connections in our defined causal structure. This is done by comparing the estimated parameters of the model on the generated data to those we defined in the structural model to generate it. Alignment of the two indicates that the causal inference method is indeed appropriate to estimate causality for (a subset of) the underlying causal relationships. If the GAN method therefore generates a synthetic dataset where the same causal inference method does not estimate the same parameters as in the generated data, this difference can be entirely attributed to the GAN method. This approach allows us to examine the causality-preservation capabilities of the GAN methods.

\subsection{Cross-sectional}
\label{section:cross-sectional-assumptions}
The first type of causal relationships are those on a cross-sectional level. Ordinary least squares (OLS) is a popular regression model to find causal effects in cross-sectional data. In this case we assume that a variable can be represented by an OLS model. The OLS model produces valid causal inference under the following assumptions:

\textbf{Assumption 1:} \textit{Linear in Parameters}

The model can be written in the form: 
\begin{equation}
    y = \beta_0 + \beta_1 x_1 + \beta_2 x_2 + ... + \beta_k x_k + \epsilon
    \label{eq: OLS_A1}
\end{equation}
\textbf{Assumption 2:} \textit{Random Sampling}\\

The sample of \textit{n} observations ${(x_{i1}, x_{i2}, ..., x_{ik}, y_i): i=1,2,...,n}$ is drawn randomly from the model.\\\\
\textbf{Assumption 3:} \textit{No Perfect Collinearity}\\

An independent variable in (\ref{eq: OLS_A1}) cannot be an exact linear combination of the other independent variables.\\\\
\textbf{Assumption 4:} \textit{Zero Conditional Mean}\\

The expected value of the error $\epsilon$ should be zero, given any values of the independent variables.

\begin{equation*}
    E(\epsilon|x_1, x_2, ..., x_k) = 0
\end{equation*}
\textbf{Assumption 5:} \textit{Homoskedasticity}\\

The error $\epsilon$ has the same variance, given any values of the independent variables.

\begin{equation*}
    Var(\epsilon|x_1, x_2, ..., x_k) = \sigma^2
\end{equation*}
\textbf{Assumption 6:} \textit{Normality}\\

The error $\epsilon$ is normally distributed a zero mean and variance $\sigma^2$ and independent of the explanatory variables $x_1, x_2, ..., x_k$.

\begin{equation*}
    \epsilon \sim Normal(0,\sigma^2)
\end{equation*}



\subsection{Time-series}
\label{section:time-series-assumptions}
In cross-sectional modelling observations have no time aspect, this changes when considering time-series models. Here we consider the popular class of linear autoregressive models. The assumptions to perform valid causal inference with these models are as follows:

\textbf{Assumption 1:} \textit{Linear in Parameters}\\

The stochastic process ${(x_{t1}, x_{t2}, ..., x_{tk}, y_t): t = 1, 2, ..., n}$ can be written in the form: 
\begin{equation}
    y_t = \beta_0 + \beta_1 x_{t1} + \beta_2 x_{t2} + ... + \beta_k x_{tk} + \epsilon_t
    \label{eq: TOLS_A1}
\end{equation}

\textbf{Assumption 2:} \textit{No Perfect Collinearity}\\

An independent variable in (\ref{eq: TOLS_A1}) cannot be an exact linear combination of the other independent variables.\\\\
\textbf{Assumption 3:} \textit{Zero Conditional Mean}\\

For each $t$, the expected value of the error $\epsilon_t$ should be zero, given any values of the independent variables.

\begin{equation*}
    E(\epsilon_t|x_{t1}, x_{t2}, ..., x_{tk}) = 0, t=1,2,...,n
\end{equation*}
\textbf{Assumption 4:} \textit{Homoskedasticity}\\

The error $\epsilon_t$ has the same variance, given any values of the independent variables.

\begin{equation*}
    Var(\epsilon|x_{t1}, x_{t2}, ..., x_{tk}) = \sigma^2, t=1,2,...,n
\end{equation*}
\textbf{Assumption 5:} \textit{No Serial Correlation}\\
Given the independent variables $x_{t1}, x_{t2}, ..., x_{tk}$, errors in two different time steps are not correlated.

\begin{equation*}
    Corr(\epsilon_s, \epsilon_t|x_{t1}, x_{t2}, ..., x_{tk}) = 0, \forall t \neq s
\end{equation*}
\textbf{Assumption 6:} \textit{Normality}\\

The error $\epsilon_t$ is normally distributed a zero mean and variance $\sigma^2$ and independent of the explanatory variables $x_{t1}, x_{t2}, ..., x_{tk}$.

\begin{equation*}
    u \sim Normal(0,\sigma^2)
\end{equation*}

Most assumptions are very similar to the previous OLS assumptions. There are two main differences. First is the absence of OLS Assumption 2 specifying observations to be randomly sampled. Under time-series assumptions observations have an order determined by the time step $t$. Second, time-series Assumption 5 is added, requiring the error term to have no serial correlation. 

In the time-series we will consider, autoregressive terms are included as well. We make an additional assumption for this autoregressive time-series to be weakly dependent, meaning the correlation between $y_t$ and $y_{t+s}$ is almost 0 for \textit{s} large enough. In other words, as the variables get farther away from each other in time, the correlation decreases.

\begin{equation*}
    y_t = \alpha y_{t-1} + \epsilon \\
\end{equation*}
In the case above of an autoregressive model lagged for one period, this assumption is satisfied if $|\alpha| < 1$.



\subsection{Structural model}
\label{section:structural-model-assumptions}
Lastly, the case remains where the whole causal structure is considered. Here, the goal is to attempt to reconstruct the full structural causal model from the data. As far as we know, no such methods exist in econometrics~\footnote{In economics, and many other fields that model complex phenomena, a structural model is defined from theory and then calibrated to data instead of trying to infer the complete model itself from the data.}. For this reason, we adopt a method from the recently developing field of causal discovery, situated mostly in the computer science literature, that tries to accomplish this task. 

Recovering the causal model from observational data is far from trivial. Recall the example above of trying to figure out the shape of the ice cube from a pool of water. As many forms of ice cubes can result in the same pool of water, many structural causal models can result in the same observational data. Therefore, picking one of all possible models is dependent on further assumptions made by each causal discovery algorithm. The general approach is to embed known features of causality, such as environment independence~\citep{arjovsky2020invariant} or acyclicity~\citep{zheng2018dags}, into the loss function that a machine learning algorithm optimises for. Even then, it is sometimes only possible to provide a set of possible structural causal models that are all equally able to generate the observational data, also called Markov equivalent. A recent trend is to extend the data to also include interventions and their outcomes~\citep{vowels2022}. This extra information can be used to exclude certain Markov equivalent models and decrease the set of potential underlying causal models.

One of the more frequently used causal discovery algorithms is LiNGAM~\citep{LiNGAM}, which assumes that the causal effects are linear, the generating causal graph is acyclic, that the distribution of the noise is non-gaussian and no unobserved confounders. The LiNGAM model can be expressed in matrix form as follows: 

\begin{equation*}
    \textbf{x} = \textbf{Bx} + \textbf{e}
\end{equation*}

with the observed variables $\textbf{x}$, the connection strength matrix $\textbf{B}$ and exogenous variables $\textbf{e}$. The condition of acyclicity allows the matrix $\textbf{B}$ to be permuted to become lower triangular with a zero-diagonal. With the additional assumption of the exogenous variables $\textbf{e}$, or in other words the noise, being non-Gaussian, the matrix $\textbf{B}$ can be uniquely identified using only the data $\textbf{x}$. This identifiability thus means that the algorithms results in a single causal graph. Different variations on this method exist like models with hidden common causes~\citep{lingam_confouders}, time-series~\citep{lingam_time} or non-linearity~\citep{lingam_nonlinear}.


In the case of Gaussian noise, only a set of Markov equivalent causal models can be estimated, while under the assumption of non-Gaussian noise this set can be reduced to one full causal model. This assumption is, however, in contrast with the assumption off Gaussian noise needed in many inference methods for valid causal inference, including the OLS and autoregressive models we discussed above.






\subsection{Generated dataset}
\label{sec: generated dataset}
We define the following model:

\begin{equation}
\begin{multlined}
    y_t = \alpha y_{t-1} + \beta_1 x_{1,t} + \beta_2 x_{2,t} + \epsilon_1 \\
    x_{1,t} = \beta_3 z_{1,t} + \beta_4 z_{2,t} + \epsilon_2 \\
    x_{2,t} = \beta_5 z_{2,t} + \epsilon_3 \\
    z_{1,t} = \epsilon_4\\
    z_{2,t} = \epsilon_5\\
\end{multlined}
\label{eq: model}
\end{equation}

\begin{figure} [h]
    \centering
    \includegraphics[width=0.5\textwidth]{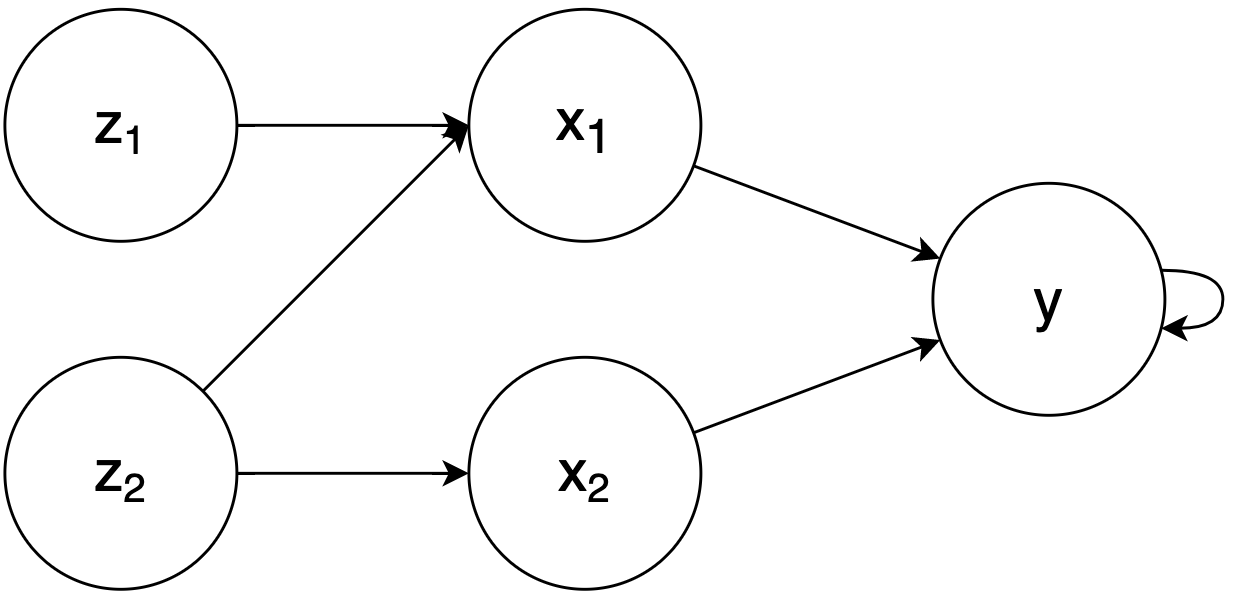}
    \caption{Full causal model of the generated dataset.}
    \label{fig: causal_model}
\end{figure}

A graphic representation of this structural model, also called the causal graph, is shown in Figure~\ref{fig: causal_model}. For the estimation of this causal structure with the different inference methods, we will always assume full observability. 

The variables $x_1$ and $x_2$ are a linear combinations of the contemporaneous values of $z_1$ and $z_2$. The underlying models for these two variables therefore meet the assumptions of the cross-sectional ordinary least squared (OLS) model. OLS should therefore be an appropriate method to estimate the causal effects of $z_1$ and $z_2$ on $x_1$ and $x_2$. We confirm this in the Results section.

For the variable $y_t$, extending the assumption on the data to allow for autocorrelation, a first order autoregressive model can infer $\alpha$ on $\beta_1$ and $\beta_2$. 

Finally a variant of LiNGAM for time-series can be used to infer the causal structural model. 

While the model was specifically chosen to contain both cross-sectional and time-series causality, it is easy to think of real-world model that follow this functional form. One example is a simple income process, where the monthly income now depends on the income last month and some contemporaneous features (e.g. employment sector, location) which in turn are distributed according to (conditionally) random distributed preferences.

%% file: chapters/generative_models.tex
\section{Generative adversarial networks}
\label{chapter/GAN}

Generative adversarial networks, or GANs, is a framework for generative machine learning first introduced by Goodfellow et al. in 2014 \citep{goodfellow2014generative}. A generative model takes a training dataset drawn from a real world distribution as input and tries to replicate this data distribution. The framework has shown great success in generating synthetic images indistinguishable from real images \citep{Photo-Realistic_GAN,brock2019large,karras2020analyzing}. While the focus has been on the improvement of the framework for image generation and manipulation, the GAN framework has recently also gathered attention for its possibilities with numerical and categorical data, like tabular and time series data.

\subsection{Framework}

A generative adversarial network consists of two competing neural networks: a generator G, which generates fake data, and a discriminator D, that is trained to discern which data is fake (made by the generator) and which data is real. The process can be described as a zero-sum game between the generator and discriminator. During the training process the generator adapts to better fool the discriminator and the discriminator in turn adapts to better detect the fake data. The resulting trained generator can then be extracted to replicate the distribution of the original data.

\begin{figure} [!b]
    \centering
    \includegraphics[width=0.9\textwidth]{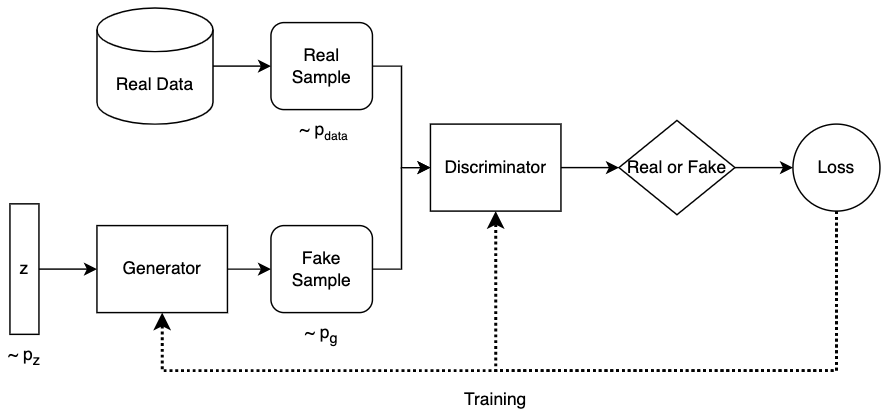}
    \caption{Generative Adversarial Networks diagram.}
    \label{fig:GAN_framework}
\end{figure}

\subsubsection{Architecture}

Figure \ref{fig:GAN_framework} shows the basic structure of a GAN. The generator G learns to map a latent space $p_z$ to a more complex distribution $p_{g}$, which is the distribution meant to mimic the real data distribution $p_{data}$. Typically, this latent space is a high-dimensional space with each variable drawn from a Gaussian distribution with a mean of zero and a standard deviation of one. The concept is thus that one can insert sample of noise ($z$) into the generator, which it will learn to map onto a sample of the distribution of the real data.
The generating function can then be described by 

\begin{equation}
    G(z) = X_{g}
\end{equation}
where $X_{g}$ are samples created by the generator. The discriminator D has the task of distinguishing the fake data $X_{g}$ from the real data $X_{data}$. The generator and discriminator are trained by playing a non-cooperative game against each other. The main aim of the generator is to produce samples which are similar to the real data. On the other hand, the main aim of the discriminator is to distinguish between fake samples from the generator and samples from the real data. The discriminator D receives both samples and tries to determine which comes from the real data distribution by assigning a probability D(x), which signifies the certainty the discriminator has in its decision. If D(x) = 1, the sample x is thought to come from $p_{data}$. On the other hand, if D(x) = 0, the discriminator judges the sample to be from $p_g$. This prediction from the discriminator and the known ground truth is then used to improve both the generator and the discriminator. During the joint training of the generator and discriminator, G will start to generate increasingly realistic samples to fool the discriminator, while the discriminator learns to better differentiate the real and fake samples. The end goal of the GAN as a whole is that the discriminator can no longer tell the difference between the generated samples $X_g$ (D(x) = 1/2) and the real data samples $X_{data}$ with the discriminator no longer able to improve itself.

\subsubsection{Loss function}

The objective function of the GAN tries to match the real data distribution $p_{data}$ with $p_g$. The original GAN \citep{goodfellow2014generative} uses two objective functions. The objective for D is to maximize the probability of assigning the correct label to both real and fake samples. This done by minimizing the negative log-likelihood for binary classification. Simultaneously G is trained to minimize log(1-D(G(z))), thus maximizing the probability of the generated samples being classified as real by the discriminator. This results in a mini-max game with objective function V(G,D):

\begin{equation}
    \min_{G} \max_{D} V(D,G) = \mathbb{E}_{x\sim p_{data}}[log D(x)] + \mathbb{E}_{z\sim p_{z}} [log(1-D(G(z)))]
    \label{eq: bceloss}
\end{equation}

The value function V(G,D) is known as the binary cross entropy function, commonly used in binary classification tasks.

\subsection{GAN extentions}
Many different variations of GANs have been proposed since its inception. In this section different relevant adaptions are presented, ordered by which level of causality they are aiming to improve.

\subsubsection{TimeGAN}
\begin{figure}
    \centering
    \includegraphics[width=\textwidth]{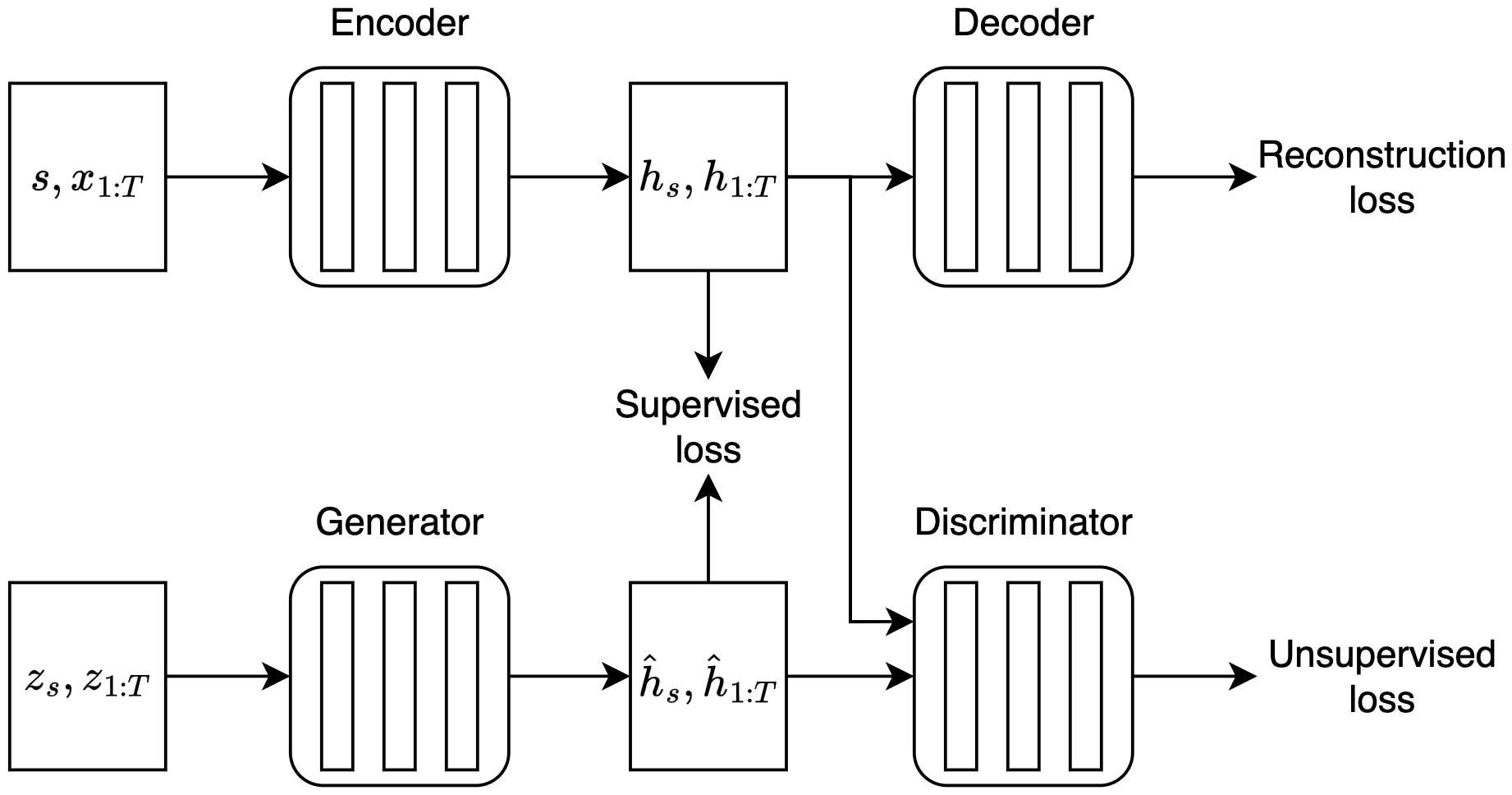}
    \caption{TimeGAN diagram}
    \label{fig:TimeGAN_orig}
\end{figure}

TimeGAN by Yoon et al. \citep{Yoon2019TimeseriesGA} is an adaptation of the original GAN framework that aims to improve the preservation of temporal dynamics for time-series data. This means that newly generated sequences should respect the original relationships between variables across time. Two main ideas are combined in the TimeGAN framework, the flexibility of the unsupervised GAN framework and a more controllable supervised autoregressive model. Figure \ref{fig:TimeGAN_orig} shows the structure of TimeGAN.

The TimeGAN framework contains the components of a generative adversarial network, as well as an auto-encoder. The latter takes as input a vector of static features, $s$, and a vector of temporal features, $x_{1:T}$. The encoder is then trained to map the feature space, which $s$ and $x_{1:T}$ belong to, to a latent space. This allows the adversarial network to learn the underlying temporal dynamics of the data via lower-dimensional representations. The output of the encoder are the latent vectors $h_s$ and $h_t$, being lower-dimensional latent codes of the input $s$ and $x_{1:T}$. In the opposite direction, the decoder takes the static and temporal latent vectors back to their feature representations. The reconstructed static and temporal features are respectively denoted as $\Tilde{s}$ and $\Tilde{x}_t$.

The other main component in the framework, the generative adversarial network, has a generator that takes as input random noise vectors and outputs latent vectors $\hat{h}_s$ and $\hat{h}_t$. The generator in this framework is autoregressive, meaning it also uses its previous outputs $\hat{h}_{1:t-1}$ for the construction of $\hat{h}_t$. A key difference with a regular GAN architecture is that the generator maps to this latent space instead of the usual feature space. Both the real latent codes $h_s$ and $h_t$ and the synthetic latent codes $\hat{h}_s$ and $\hat{h}_t$ are received by the discriminator, which has the task to classify these codes as either real or fake.

The resulting framework has three loss functions. First, the reconstruction loss. This loss is linked to the auto-encoder component of the framework, quantifying the difference between original features $s$, $x_{t}$ and the reconstructed features $\Tilde{s}$ and $\Tilde{x}_t$.

\begin{equation}
    \mathcal{L}_R = \mathbb{E}_{s,x_{1:T}\sim p}[||s-\Tilde{s}||_2 + \sum_t ||x_t-\Tilde{x}_t||_2]
\end{equation}
Second, the unsupervised loss is the same type of loss used in the original GAN framework, maximising (discriminator) or minimising (generator) the likelihood of providing correct classifications. Notations $y$ and $\hat{y}$ denote classifications by the discriminator as respectively real or synthetic data.

\begin{equation}
    \mathcal{L}_U = \mathbb{E}_{s,x_{1:T}\sim p} [\log y_s + \sum_t \log y_t] + \mathbb{E}_{s,x_{1:T}\sim \hat{p}} [\log (1-\hat{y}_s) + \sum_t \log (1- \hat{y}_t)]
\end{equation}
Lastly, the supervised loss is introduced. The addition of this loss is motivated by the idea that the regular feedback from the discriminator, the unsupervised loss, may be insufficient incentive for the generator to capture the step-wise conditional distributions in the data. To calculate this loss, the autoregressive generator $g$ uses the real latent codes $h_s$ and $h_{t-1}$ instead of the synthetic $\hat{h}_s$ and $\hat{h}_{t-1}$ to generate $\hat{h}_t$, or $g(h_s, h_{t-1}, z_t)$, as shown in (\ref{eq: superv_loss}).

\begin{equation}
    \mathcal{L}_S = \mathbb{E}_{s,x_{1:T}\sim p} [\sum_t ||h_t - g(h_s, h_{t-1}, z_t)||_2] \label{eq: superv_loss}
\end{equation}

A linear combination of $\mathcal{L}_U$ and $\mathcal{L}_S$ is used to train the generator and the discriminator. $\mathcal{L}_U$ guides the generator to create realistic sequence, while $\mathcal{L}_S$ uses ground-truth targets to ensure that the stepwise transitions are similar. To train the autoencoder components, the encoder and the decoder, a linear combination of $\mathcal{L}_R$ and $\mathcal{L}_S$ is used. By combining the different objectives, TimeGAN is trained to simultaneously encode feature vectors, generate latent codes for these feature vectors, and iterate across time.


\subsubsection{CausalGAN}
\label{sec: causalGAN}
CausalGAN is a generative adversarial framework proposed by Kocaoglu et al.~\citep{kocaoglu2017causalgan}. CausalGAN is an implicit causal generative model that replicates data constraint to a given causal graph. Implicit generative models, which the original GAN model is part of, can sample from a probability distribution, without the ability to provide likelihoods for the samples~\citep{implicit_generative}. Causal implicit generative models can not only sample from a probability distribution but also from conditional and interventional distributions, which causal graphs embeds.

\begin{figure}
     \centering
     \begin{subfigure}[b]{0.3\textwidth}
         \centering
         \includegraphics[width=\textwidth]{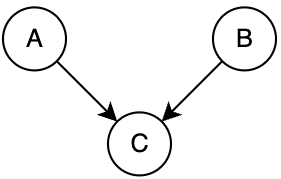}
         \caption{Causal graph $A \xrightarrow{} C \xleftarrow{} B$}
         \label{fig: Causal_GAN_graph}
     \end{subfigure}
     \hfill
     \begin{subfigure}[b]{0.65\textwidth}
         \centering
         \includegraphics[width=\textwidth]{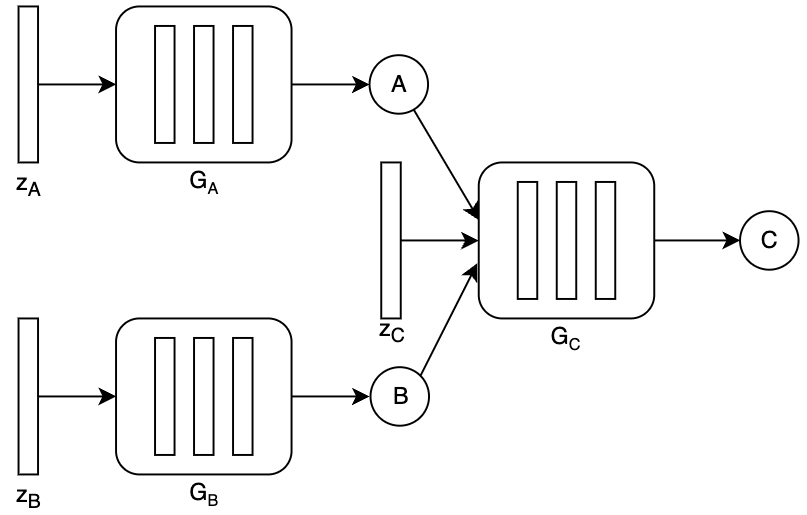}
         \caption{Generator architecture for $A \xrightarrow{} C \xleftarrow{} B$}
         \label{fig: Causal_GAN_generator}
     \end{subfigure}
        \caption{}
        \label{fig:Causal_GAN_total}
\end{figure}
Consider a simple causal graph, $A \xrightarrow{} C \xleftarrow{} B$, as depicted in Figure \ref{fig: Causal_GAN_graph}. The parent nodes, \textit{A} and \textit{B} are assumed to have no other variables influencing their distribution and can be written as $A = G_A(Z_A)$ and $B = G_B(Z_B)$, where $Z_*$ is some chosen noise distribution (e.g. Gaussian), and $G_*$ is a function mapping this distribution to the distribution of the variable. The variable $C$ has two parent nodes and can be written as $C = G_C(A,B,Z_C)$, being a function of both $A$ and $B$, as well as a chosen distribution. This representation is similar to how the generator of the original GAN framework is structured. Figure \ref{fig: Causal_GAN_generator} shows how a generator can be constructed to represent a given causal graph. For each variable a feedforward neural network is used represent functions $G_*$, resulting in a larger generator network consisting of linked individual generators. By building in the causal graph into the generator, it will constrain the data generation data to the actual causal model and not only reproduce joint probabilities but also the causal relationships. For the implementation of CausalGAN in this paper Causal-TGAN~\citep{causal_tgan} is used. This version uses the same core idea as CausalGAN, with some added adjustments for tabular data.

The downside is of course that both data and the relevant causal graph needs to be known to train and use the generator. To this end, we use a causal discovery method, here the standard- and time-variant of LiNGAM to provide us with the causal graph of the data.

%% file: chapters/results.tex
\section{Results}

Consider the model described in Section \ref{sec: generated dataset} with the following parameters:

\begin{table}[h]
\centering
\begin{tabular}{l|l}
Parameter & Value \\\hline
$\alpha$       &  0.5   \\
$\beta_1, \beta_2, \beta_3, \beta_4, \beta_5$   &    1   \\
$\sigma_1, \sigma_2$     &    1   \\

\end{tabular}
\end{table}
where $\epsilon_* \sim N(0,0.5)$. From this model we sample $10,000$ observations to use for further experiments. These observations will further be referred to as generated data and will be used to both train the different GAN models and give baseline values for estimated parameters. The experiments assume a perfect scenario where the model is known. Each experiment is done, in its entirety, $10$ times and reported results show averages and standard deviations over these $10$ runs.

\subsection{GAN}
\label{section:GAN}
First, we train a standard GAN with the generated data. From this GAN, we generate $10,000$ samples to preserve the statistical power of our inference results. These latter samples will be referred to as the synthetic data. The first model we fit on both datasets is OLS for the following variables:

\begin{equation*}
\begin{multlined}
    x_{1} = \beta_3 z_{1} + \beta_4 z_{2} + \epsilon_2 \\
    x_{2} = \beta_5 z_{2} + \epsilon_3 \\
\end{multlined}
\end{equation*}

The resulting parameters can be seen in Table~\ref{tab: results GAN}. The results show that on a cross-sectional level, with the underlying model meeting the assumptions in~\ref{section:cross-sectional-assumptions}, the GAN methodology can replicate data with similar causal relationships. While on average the causal relationships detected in the synthetic data are less accurate than the causal relationships in the generated data, the results are not significantly different from the true parameters.

While the data the GAN is trained on is time-ordered, the synthetic data produced by the GAN is sampled randomly, without any notion of time. So, as expected, when running an autoregressive model on the $y$ variable in our model, it does not find any time-correlation ($\alpha$ coefficient for $y$) in the synthetic data. Interestingly, it does capture the cross-sectional relationships for $y$ ($\beta_1$ and $\beta_2$).


\begin{table}[h]
\centering
\begin{tabular}{ll|lllll}
           Model & Par. &  Real & GAN & TimeGAN & CausalGAN\\ \hline
OLS & $\beta_3$ &  0.9990 $\pm$ 0.0051  & 1.0209 $\pm$ 0.0715 & 0.3762 $\pm$ 0.4320 & 0.9869 $\pm$ 0.1087 \\
    & $\beta_4$ &  1.0017 $\pm$ 0.0052  & 1.0797 $\pm$ 0.1272 & 1.2249 $\pm$ 0.3362 & 0.9666 $\pm$ 0.1029 \\
    & $\beta_5$ &  0.9996 $\pm$ 0.0057 &  1.0157 $\pm$ 0.1266 & 1.1066 $\pm$ 0.0179 & 1.0006 $\pm$ 0.1625 \\\hline 
TS     & $\alpha$ &  0.5004 $\pm$ 0.0011  & 0.0030 $\pm$ 0.0020 & 0.0233 $\pm$ 0.1331 & 0.0011 $\pm$ 0.0064\\
       & $\beta_1$ &  0.9993 $\pm$ 0.0040  & 1.0007 $\pm$ 0.1773 & 1.0597 $\pm$ 1.5236 & 0.9635 $\pm$ 0.1896\\
       & $\beta_2$ &  0.9982 $\pm$ 0.0045  & 1.1439 $\pm$ 0.1682 & 0.8796 $\pm$ 2.0436 & 0.9927 $\pm$ 0.2035\\
\end{tabular}
\caption{Results for all GANs}
\label{tab: results GAN}
\end{table}

\subsection{TimeGAN}
Next, TimeGAN is trained on the generated dataset, after which we again sample $10,000$ datapoints for a new synthetic dataset. Note that the sampled datapoints are now ordered in time instead of randomly sampled as in section~\ref{section:GAN}. 

As can be seen in Table~\ref{tab: results GAN}, the synthetic data produced by TimeGAN does not properly maintain causal relationships, neither on a cross-sectional level nor over time. The results are far from what would be expected and also vary significantly from run to run, resulting in a higher standard deviations in the results. This is likely due to there being no auto-correlation in the variables outside of $y$, and TimeGAN attempting to find time dependent structure where none exists. To confirm this, we also consider the following alternate causal structure, where all variables have some sort of time-dependence (direct or indirect). 

\begin{equation}
\begin{multlined}
    y_t = \alpha y_{t-1} + \beta_1 x_{1,t} + \beta_2 x_{2,t} + \epsilon_1 \\
    x_{1,t} = \beta_3 z_{1,t} + \beta_4 z_{2,t} + \epsilon_2 \\
    x_{2,t} = \beta_5 z_{2,t} + \epsilon_3 \\
    z_{1,t} = z_{1,t-1} + \epsilon_4\\
    z_{2,t} = z_{2,t-1} + \epsilon_5\\
\end{multlined}
\label{eq: model alt}
\end{equation}

Table~\ref{tab: results TimeGAN alt} shows the results for TimeGAN in the case of the alternative structure. In this case TimeGAN is able to accurately capture the causal relationships on a cross-sectional level ($\beta_3$, $\beta_4$, $\beta_5$) but still fails to capture the structure in $y$ ($\alpha$, $\beta_1$ and $\beta_2$). However, it does not seem like the model completely missed the mark. When we look at the original formulation for $y$, with the chosen parameters for the experiment, it can be rewritten as follows:

\begin{equation*}
    y_t = 0.5 y_{t-1} + x_{1,t} + x_{2,t} + \epsilon_t
\end{equation*}
\begin{equation*}
    y_t = 0.25 y_{t-2} + (x_{1,t} + 0.5 x_{1,t-1}) + (x_{2,t} + 0.5 x_{2,t-1}) \\
    + (\epsilon_t + 0.5 \epsilon_{t-1}) \\
\end{equation*}

\begin{equation*}
\begin{multlined}
    y_t = 0.125 y_{t-3} + (x_{1,t} + 0.5 x_{1,t-1} + 0.25 x_{1,t-2}) \ + \\
    (x_{2,t} + 0.5 x_{2,t-1} + 0.25 x_{2,t-2}) + (\epsilon_t + 0.5 \epsilon_{t-1} + 0.25 \epsilon_{t-2})   
\end{multlined}
\end{equation*}\\
This decomposition of $y$ can be continued further until the autoregressive part for y is negligible. Now, if the change in $x_1$ and $x_2$ in each time step is limited and thus $x_{1,t} \approx x_{1,t-1}$ and $x_{2,t} \approx x_{2,t-1}$, as is the case here due to the stationarity of $y$, and using $\sum_{n=0} (\frac{1}{2})^n = 2$, we can write:
\begin{equation*}
    y_t \approx 2x_{1,t} + 2x_{2,t} + \epsilon
\end{equation*}\\
with $\epsilon \sim N(0, \frac{4}{3})$. The results shown in Table~\ref{tab: results TimeGAN alt} thus suggest that TimeGAN has learned this smaller representation of $y$, using only $x_1$ and $x_2$, that results in the same expected values of $y$ over time. This representation, however, does not represent the actual causal model underlying $y$.

\begin{table}[]
\centering
\begin{tabular}{ll|lll}
           Model & Parameter &  Real & TimeGAN\\ \hline
OLS & $\beta_3$ &  0.9999 $\pm$ 0.0002  & 0.9967 $\pm$ 0.0247 \\
    & $\beta_4$ &  1.0000 $\pm$ 0.0001  & 1.0049 $\pm$ 0.0219 \\
    & $\beta_5$ &  0.9999 $\pm$ 0.0001 &  1.0005 $\pm$ 0.0024 \\\hline 
TS     & $\alpha$ &  0.4999 $\pm$ 0.0008  & -0.0128 $\pm$ 0.0207\\
       & $\beta_1$ &  1.0000 $\pm$ 0.0016  & 2.0719 $\pm$ 0.0680\\
       & $\beta_2$ &  1.0000 $\pm$ 0.0016  & 2.0002 $\pm$ 0.1550\\
\end{tabular}
\caption{Result for TimeGAN on the second model}
\label{tab: results TimeGAN alt}
\end{table}

\subsection{CausalGAN}
Lastly, the full structural causal model is considered. Here, a model can not be directly trained to the data since no such method exists as far as the authors are aware. A two-step approach is taken where first the causal structure is identified with LiNGAM. This extracted structure is then compared to our data generating model (Eq.~\ref{eq: model}) to check if LiNGAM is an appropriate and efficient causal discovery method for our case. Then CausalGAN is used to generate data that follows this structure. Lastly, LiNGAM is applied to the synthetic data and its output is compared to the causal structure retrieved from the generated data.

As noted before, LiNGAM uses the assumption of non-Gaussian noise, which is incorrect for model~(\ref{eq: model}) used previously in this section. To start from a correct causal structure for this experiment, we adjust the distribution of the noise our data structure~(\ref{eq: model}) to be uniformly distributed, $\epsilon_* \sim U(-1,1)$. Under these conditions the time-variant of LiNGAM is able to find the underlying causal model correctly. However, CausalGAN is not equipped to deal with time-series, so we are forced to only consider the cross-sectional causal relations here.

Table~\ref{tab: results CausalGAN} shows all causal relationships detected by LiNGAM in both the generated dataset and the synthetic dataset produced by CausalGAN. Additionally, we show the causal relationships detected in synthetic data from a basic GAN trained on the real data. For this one representative example is chosen since the use of means and standard deviations give warped representations of the results. The synthetic data sampled from CausalGAN consistently maintains causal relationships relatively well. Some deterioration can be seen, as well as introducing small additional causal effects. The basic GAN framework is however not capable of retaining the causal relationships when the whole causal structure is considered. Causal discovery on the synthetic data of the basic GAN gives varying results even when performed multiple times on one synthetic dataset. None of the resulting graphs are close to the original causal graph. This shows that adding the additional information of the (correct) underlying causal graph through the CausalGAN model does help maintaining the causal structure.



\begin{table}[h]
\centering
\begin{tabular}{l|lll}
            Causal effect & Real & CausalGAN & GAN\\ \hline
            $z_{1} \xrightarrow{} x_{1}$ & 1.00 & 0.93 & 1.03\\
            $z_{2} \xrightarrow{} x_{1}$ & 1.01 & 0.80 & \textbf{1.07}\\
            $z_{2} \xrightarrow{} x_{2}$ & 0.99 & 0.83 & \textbf{0.16}\\
            $x_{1} \xrightarrow{} y$ & 1.02 & 1.04 & \textbf{0.14}\\
            $x_{2} \xrightarrow{} y$ & 1.01 & 1.00 & \textbf{0.39}\\\hline
            $z_{1} \xrightarrow{} z_{2}$ & 0 & 0 & -1.11\\
            $z_{1} \xrightarrow{} x_{2}$ & 0 & 0 & -0.47\\
            $z_{1} \xrightarrow{} y$ & 0 & 0 & 0.86\\
            $z_{2} \xrightarrow{} y$ & 0 & 0.14 & \textbf{-0.10}\\
            $x_{2} \xrightarrow{} x_{1}$ & 0 & 0.14 & 0.65\\

\end{tabular}
\caption{Causal effects detected by LiNGAM on both the generated dataset and the synthetic dataset generated by CausalGAN. The table contains all significant causal effects ($> 0.1$). Causal effects of less significance ($< 0.1$) are simplified to 0. \textbf{Bold} number indicate that the causal effect is reversed}
\label{tab: results CausalGAN}
\end{table}

%% file: chapters/applications.tex
\section{Real world challenges}
In our tests of the causality replicating capabilities of GANs, we have purposely abstracted away from many of the additional challenge that come with working with real-world data. In this section we address three of the most important challenges and give an overview of the variations on the GAN framework that have been proposed to tackle them. 

\subsection{Privacy}
Privacy concerns are one of the main drivers for the recent rise in interest in synthetic data. While in general synthetic data is sampled from a reconstruction of the distribution of the original data, fear of replicating real samples due to overfitting remain~\citep{hybrid_overfitting, Feng2022}. Membership inference attacks also form a common concern in the field of privacy~\citep{hayes2017, Chen_2020}. These attacks leverage the fact that machine learning models generally perform better on the data it was trained on to reconstruct the training data.

These concerns have sparked the search for GAN variants that give certain privacy guarantees. One such guarantee is differential privacy. An algorithm is differentially private if an observer seeing the output can not tell if a particular datapoint was used in the computation. In the case where the observer has access to the generated samples but not the generator, recent work has shown that the base form of GAN has some privacy guarantees in terms of both differential privacy and robustness to membership inference attacks~\citep{Lin_2022}. These guarantees get stronger for larger training datasets. If additionally the generator is available, several differential privacy GANs have been proposed, such as DPGAN~\citep{DPGAN}, PPGAN~\citep{PPGAN} and PATE-GAN~\citep{yoon2018pategan}. 

Privacy guarantees, however, come at the price of replication quality since you in some form or another adding noise to the data by limiting the impact a training sample can have on the model, even though it might be highly informative~\citep{Huang_2017_GAP, Lin_2020}. 

\subsection{Fairness}
Machine learning has an increasingly large impact on current day decision making, scaling decisions made on a micro-scale to a macro-scale in an often opaque manner. This trend has raised concerns about building in, or scaling up biases in decisions. Fairness in machine learning is a recently growing area of research that studies how to ensure that such biases and model inaccuracies do not lead to discriminatory models on the basis of sensitive attributes such as gender or ethnicity. Using synthetic data can help by debiasing the data before it even gets used for further analysis. In such a framework a generative model is trained on unfair data to generate synthetic fair data.

A first challenge to fairness is defining what it actually is, which is often highly dependent on the context of the business decisions that is being made with the model. One often used interpretation is that certain features, also called protected or sensitive features (e.g. gender, ethnicity), should not have any impact on the outcome of the model. This orthogonalisation of the model outcome and the protected features comes with two major challenges. First, it requires outside definition of what the protected features are. Second, if you want to rid observational data of such biases, it is not enough to just delete the features, you need to know the relevant causal structures to exclude both the direct and indirect impact the protected attribute has on the outcome~\citep{NIPS2017_a486cd07, fairness_causal}. Otherwise the model can just learn the protected features by using different proxies which are correlated to them~\citep{DECAF}. CFGAN~\citep{causal_fairness_gan} and DECAF~\citep{DECAF} are two methods to generate fair data that are rooted in this approach to fairness. Both methods therefore require a causal graph as additional input, something we saw in our results is not generally feasible with current causal discovery methods.

FairGAN~\citep{FairGAN} and Fairness GAN~\citep{Fairness_GAN} have also been suggested for the purpose of generating fair data. FairGAN uses an additional discriminator on top of the classical GAN architecture to determine whether samples are from the protected or unprotected group. Fairness GAN uses an added loss function that encourages demographic parity. Demographic parity is satisfied if the decisions made from the data are not dependent on a given sensitive attribute. This requires a specification of the explanatory variables \textit{x}, the target variables \textit{y} and the sensitive variables \textit{s}, where \textit{y} does not need specification in other methods. FairGAN is applied to low-dimensional structured data, while Fairness GAN is applied to high dimensional image data. 

\subsection{Tabular data}
Tabular data is data that contains both discrete and continuous columns and is one of the most commonly encountered data formats in both business and research~\citep{TGAN}. Tabular data, and especially the discrete features within them are challenging for GAN methods since the continuous functions used in neural nets are ill-equipped to fit the non-continuous distributions of discrete variables. 

The generator of a regular GAN cannot generate discrete samples because the generator is trained by the loss from the discriminator via backpropagation \citep{goodfellow2014generative}. To tackle this problem, MedGAN~\citep{choi2018generating} adds an autoencoder model to the regular GAN framework to generate high-dimensional discrete variables. Both TGAN~\citep{TGAN} and TableGAN~\citep{park2018data} look to improve the performance on the continuous distributions as well. TGAN clusters numerical variables to deal with the multi-modal distribution for continuous features and adjusts the loss function to effectively generate discrete features. TableGAN uses a classifier neural network to predict synthetic records’ labels to improve consistency in generated records. An additional loss, information loss, is introduced as well. This loss is the difference in key statistical values of both the real and synthetic data. In the paper the mean and standard deviation are used as key statistical properties.

Besides the mix of continuous and discrete columns, the distributions of data often differs from the standard Gaussian-like distributions found in typical generative applications like image generation. To this end CTGAN~\citep{CTGAN} addresses additional concerns about non-Gaussian and multi-modal distributions, and imbalanced categorical columns. CTAB-GAN~\citep{CTAB-GAN} looks further into these issues and tackles data imbalance and long-tail distributions. The previously mentioned Causal-TGAN~\citep{causal_tgan} combines ideas of CTGAN and CausalGAN~\citep{kocaoglu2017causalgan} to leverage knowledge about the causal structure for a better performance.

%% file: chapters/conclusion.tex
\section{Conclusion}
Data has become a driving force in both business, research, and policy. And rightfully so if we see how increased access to data has furthered our ability to understand and support decision making in complex environments. While some fields are just collecting more and more data in labs or in nature, most of the decision making that occurs in business is in regards to actual human beings. This rightfully raises concerns about privacy and ethics. Should companies just be allowed to collect, buy, sell, and share more and more data on the behaviour and features of actual human beings just for the sake of making better business-decisions? The answer is obviously no, and regulatory bodies are acting accordingly by setting in place boundaries on what is and is not allowed in regards to data on individuals. 

How do we balance the benefits of increased accuracy and understanding with the privacy and ethics concerns that both come with having more data? One solution that has gained a lot of traction is synthetic data, which are data sampled from generative methods that are meant to replicate high-dimensional distributions of data. After all, the improvements in modelling complex phenomena come from sufficient coverage of the high-dimensional distribution of relevant features, and not from knowing someone's exact name or address. So if we could generate data with the same distribution as the original, but not containing any identifiable features as well as different enough in exact values such that no individual could be uniquely linked to one sample from the data, we could have all the benefits without introducing risks to privacy or ethics.

While this is true for predictive models, that solely map correlations to an outcome, many decisions intend an intervention to influence the outcome. The difference lies in that the former asks an observational question: ``If I observe X, what will Y be?'' and the latter asks an interventional question: ``If I do X, how will the outcome Y change''. Apart from a group of so-called \emph{policy prediction problems}~\citep{kleinberg2015}, which only require a prediction to make a decision, the latter requires causal inference. Once we enter into this territory, it no longer just matters that the synthetic data has the right distribution, but also that it was generated with the correct underlying causal relationships. and because there can exist multiple underlying structures that generate the same distribution, there are no intrinsic guarantees that current generative modelling methods converge on the correct one.

We evaluate the causal replication capabilities of the generative modelling techniques that are typically used for synthetic data. As far as we know, we are the first to do so with a focus on causality. We find that in the case where the assumptions are met that make correlation equal causation, causal inference on the real and synthetic data yield the same results only if the simplest model that can generate the distribution of the features equals the real one. This points at the principle of occam's razor, that is the foundation for regularisation in machine learning to counter overfitting, is actually working against us in the case where we want to replicate causal relationships.

When nothing is known about the causal structure, and the analyst can thus not easily construct a functional form to test with classic causal inference methods like OLS, causal discovery can be used. Causal discovery tries to find the complete causal structure in observational data, which can then be used as input for a generative model that can generate synthetic data explicitly according to the causal structure. We find that, while this works in simple cases (e.g. in the case of cross-sectional correlation with non-gaussian noise), the necessary assumptions on both the causal discovery and generation side seem too restrictive to be widely applicable in real-world contexts.

A path forward seems to be to augment the observational data fed to the GAN models with additional information such as knowledge on different environment in which the data was collected or interventional data from experiments~\citep{Scholkopf2021}. While this can present a way forward for many fields, it is often not applicable in the context of businesses related to people's finances or health. 

Organisations that want to improve their decision making by leveraging synthetic data should thus be careful about what the current state-of-the-art is actually capable of.